# Multi-Level Feature Fusion Mechanism for Single Image Super-Resolution

Jiawen Lyn
School of Computer Science and Statistics
Trinity College Dublin
Dublin, Ireland
linj1@tcd.ie

***Abstract*—Convolution neural network (CNN) has been widely used in Single Image Super Resolution (SISR) so that SISR has been a great success recently. As the network deepens, the learning ability of network becomes more and more powerful. However, most SISR methods based on CNN do not make full use of hierarchical feature and the learning ability of network. These features cannot be extracted directly by subsequent layers, so the previous layer hierarchical information has little impact on the output and performance of subsequent layers relatively poor. To solve above problem, a novel Multi-Level Feature Fusion network (MLRN) is proposed, which can take full use of global intermediate features. We also introduce Feature Skip Fusion Block (FSFblock) as basic module. Each block can be extracted directly to the raw multi-scale feature and fusion multi-level feature, then learn feature spatial correlation. The correlation among the features of the holistic approach leads to a continuous global memory of information mechanism. Extensive experiments on public datasets show that the method proposed by MLRN can be implemented, which is favorable performance for the most advanced methods.**

## I. Introduction

Image super-resolution refers to reconstructing a low-resolution image into a high-resolution image. High resolution means the pixels in images are denser and can display more texture detailed features. These details are very useful in practical applications, such as satellite imaging, medicine, etc. In areas such as imaging and high-resolution images can better identify targets.

Prior to convolutional neural networks, image super-resolution techniques have been divided into interpolation-based methods [1], model-based reconstruction methods [2], learning-based methods [3], and interpolation-based algorithms that mainly uses low-resolution images. The local area known pixels infer the unknown pixel values in the high-resolution image. Common interpolation algorithms are bicubic interpolation, nearest neighbor interpolation, and bilinear interpolation. This type of method is simple in principle and fast in speed, but when the magnification (3 times or 4 times) is large, the image as a whole becomes blurred, and sharp edge features are smoothed without showing detailed features. The reconstruction model-based algorithm constrains the consistency between low-resolution and high-resolution images by constructing models, generally existing in known prior conditions, for example, Gaussian distribution. Because not all the gradients of the image are in a prior condition, the image reconstructed by this method has poor quality and edges are seriously blurred.

Due to the limitations of traditional methods and the improvement of the deep learning theory in recent years, convolutional neural networks have achieved remarkable achievements in the fields of image classification and target positioning. Therefore, many algorithms based on convolutional neural networks have appeared in image super-resolution technology. Dong et al. [4] first introduced the use of convolutional neural networks to implement image super-resolution (SRCNN) in 2014. The SRCNN network consists of three layers, namely feature extraction layer, nonlinear mapping layer and reconstruction layer. Due to the powerful learning power of convolutional neural networks, SRCNN is significantly better than traditional algorithms. However, SRCNN is to enlarge the low- resolution image by double-cubic interpolation, using it as the input of the network. This method makes the calculation of network large with a lot of time cost. According to this, Dong et al [5] proposed FSRCNN in 2016, the input of network was changed to low-resolution image, and the depth of the network was increased. Finally, the low-dimensional high-dimensional feature map was transformed into a high one by deconvolution so as to distinguish images. FSRCNN is superior to SRCNN in the effect and speed of over-score. In 2016, Shi et al. invented the sub-pixel convolution [6] to convert low-dimensional high-dimensional features into high-resolution images. Because of the large number of complementary zero operations in deconvolution, sub-pixel convolution is better than the deconvolution reconstruction. Both FSRCNN and Sub-Pixel CNN use a convolutional neural network with a small number of layers. Later, Kim et al. proposed the VDSR model [7] in 2016. It contains 20 layers of residual blocks, which outperformed the FSRCNN network in super-resolution image quality. Tai et al. [8] proposed a deep network called MemNet [8] consisting of cascaded memory blocks which can fuse global features for better vision result.

However, there are still some problems with methods mentioned above. Although FSRCNN and ESPCN improve the network training speed, there are less convolution layers. ESPCN has only three layers of convolution operations which is unable to fully extract more detailed features of the image, and the network model is incorrect. The boundary uses zero padding so that the output image can be reduced in size after the convolution operation, and the edge information of the image cannot be fully utilized. Although the VDSR deepens the network depth, its input image needs to be preprocessed by bicubic interpolation, making the amount of calculation increase When the calculation speed slows down, and the network is difficult to converge. Besides, MemNet interpolates LR images as preprocessing so as to get the HR images of the same size and not straightway extracted original low-resolution images features.

In order to solve above problems, we propose a new feature skipping fusion Network (MLRN). The MLRN contains a shallow feature extractor and a deep feature extractor and reconstruction layer. The details of shallow feature extractor will be introduced in the Section 3. The depth feature extractor consists of multiple cascaded Feature Skipping Fusion Block (FSFblocks). First, we use shallow feature extractor to obtain shallow image features. Second, FSFblocks extracts different scale feature of image which is considered to local multi-scale features. Third, the output of each RSFblock is concatenated for global feature fusion. As a result, the combination of local multi-scale features and global features is upscaled. Finally, the upscaled feature via a correcting layer to a correct feature. In addition, we introduced a convolutional layer with a 1×1 kernel for global feature fusion.

In summary, this work includes following three main contributions:

- A deep end-to-end network Feature Skipping Fusion Network (MLRN) is introduced to solve image super-resolution reconstruction problem of different scale factor. The network have the ability that learn multi-scale feature, fusion local multi-level feature, global feature from LR images and directly reconstruct HR images. Extensive experiments on public datasets that demonstrate the superiority of our MLRN.
- A Feature Skipping Fusion Block (FSFblock) is proposed for MLRN, which establishes multi-level connection among multi-scale features. FSFblock learns the multi-level feature and spatial correlation to extract deeper features.
- We propose a simple architecture named Multi-Level Feature Fusion (MLFF) for super resolution image reconstruction. It can be easily extended to any deep learning network.

## II. Methods and Network Structures

### A. Authors and Affiliations

As shown in Fig 1 our Multi-Level Feature Fusion network (MLRN) architecture consists of four parts: a shallow feature extraction block (SFblock), dense feature skip fusion blocks (FSFblocks), reconstruction layer, and a deep feature extraction block (DFB), as shown in Fig 1. MLRN is to reconstruct a super-resolution $I^{LR}$ from a low-resolution image $I^{LR}$. The $I^{LR}$ is obtained by the bicubic operation from $I^{HR}$. As for an image with $C$ color channels, we denote the $I^{LR}$ with the shape of $W \times H \times C$ and the $I^{HR}$, $I^{SR}$ with $rW \times rH \times C$, where $C = 3$, representing the RGB channel. $r$ represents the upscaling factor. We use a convolution layer to extract the coarse features from the $I^{LR}$:

$$F_{-1} = H_{extract}(I^{LR}) = W_0 \times I^{LR} \tag{1}$$

where $H_{extract}$ denotes the coarse feature extraction function, the weight of the convolution layer is $W_0$ and $F_{-1}$ is the coarse feature from $I^{LR}$. $F_{-1}$ is used for SFblock based on shallow feature extraction, which thus produces the shallow feature as:

$$F_0 = H_{SF}(F_1) \tag{2}$$

where $f_{SF}$ represents the SFblock based on shallow feature extraction module, which consists of two convolution layers and shares residual skip. SFblock obtains more useful features for training. Then the shallow feature $F_0$ is used for FSFblocks based on deep feature extraction, thus producing the deep feature as:

$$F_d = H_{FSFB,d}(F_{d-1}) \tag{3}$$
$$= H_{FSFB,d}(f_{FSFB,d-1}(\cdots((\cdots(H_{FSFB,1}(F_0))\cdots))$$

where $f_{FSFB,d}$ represents the operations of the d-th FSFblock based deep feature extraction module, which consists of several feature skip fusion blocks. FSFblock consist of multi-scale convolution layer and complex feature fusion. After extracting the multi-level features with a set of FSFblocks, we further conduct the global feature fusion (GFF), which includes global feature fusion (GFF) and residual skip connection (RSC). GFF makes full use of multi-level feature from all the preceding FSFblocks and can be represents as:

$$F_{DF} = H_{GFF}(F_1, F_2, F_3, \cdots, F_d) \tag{4}$$

where $F_{DF}$ is the output of $H_{GFF}$ by utilizing dense feature fusion. Hence, our proposed network has high complexity and can obtain a very deep feature. Then the deep feature $F_{DF}$ is upscaled via the upscale module via:

$$F_{\uparrow} = f_{\uparrow}(F_{DF}) \tag{5}$$

where $f_{\uparrow}$ and $F_{\uparrow}$ are upscale module and upscaled feature respectively. The ways to act upscale part have some choices, such as transposed convolution [5] and sub-pixel convolution [6]. The way of embedding upscaling feature in the last several layers obtains a good trade-off between computational burden and performance so that it is preferred to be used in recent CNN-based SR models [5, 9,10]. The upscaled feature $F_{\uparrow}$ is then mapped into SR image via a feature reconstruction layer for correcting feature.

$$I^{SR} = f_C(F_{\uparrow}) = f_{FSFN}(I^{LR}) \tag{6}$$

where $f_C$, $f_{\uparrow}$ and $f_{FSFN}$ are the reconstruction layer, upscaled layer and the function of SAN respectively.

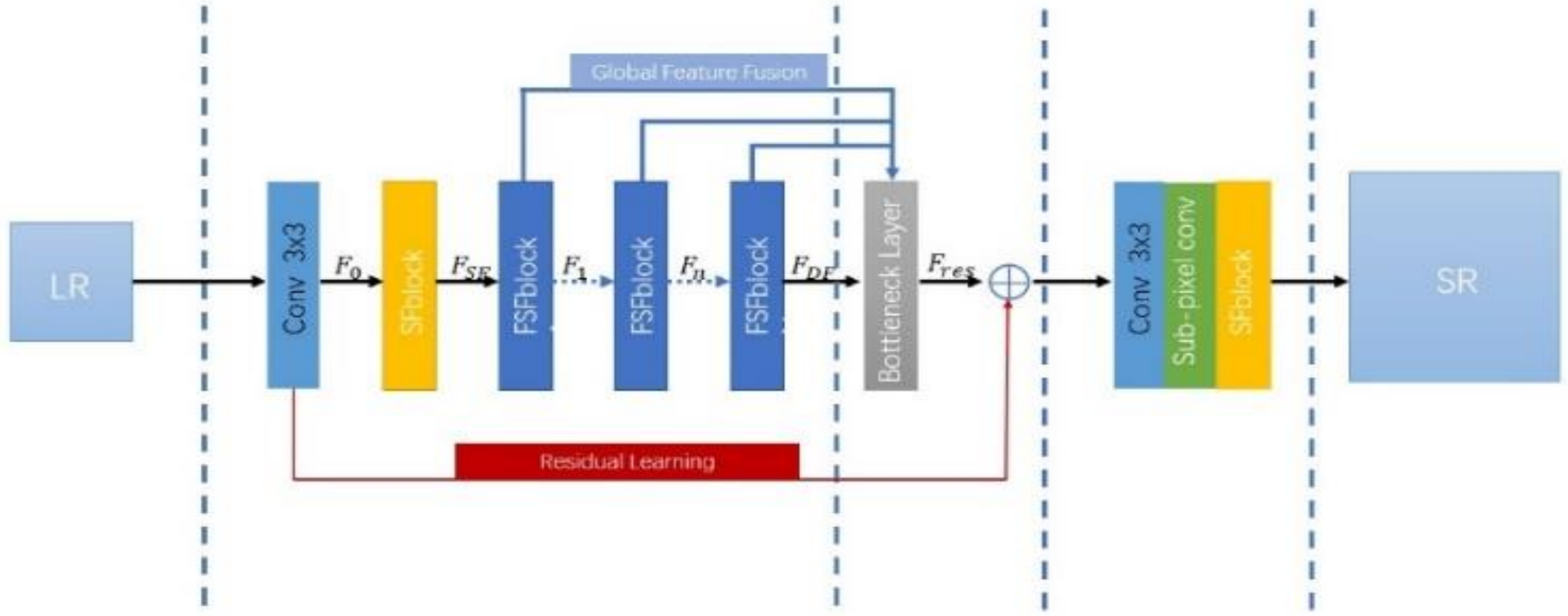

Fig. 1. The basic architecture of our proposed Multi-Level Feature Fusion network (MLRN). Red arrow and black arrows mean residual information flow and global information flow respectively. Blue arrows mean the global feature fusion information flow.

## B. *Shallow Feature Block*

After extracting coarse feature with a convolution layer, we further propose shallow feature block (SFblock) to extract shallow feature and correct the feature in the tail of the network. Our SFblock consists of several 1 x 1 convolution layers and local residual learning. The relevant details are shown in Fig 2.

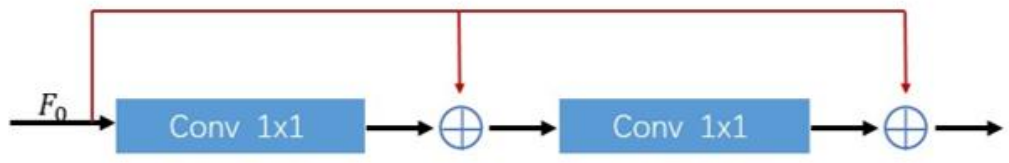

Fig. 2. Shallow feature extract block, which consists of two 1 x 1 convolution layers and local residual learning. Red arrows and black arrows mean residual information flow and global information flow respectively.

## C. *Feature Skip Fusion Block*

Now we provide details on our proposed feature skip fusion block (FSFblock) in Figure 3. Our FSFblocks contain four main parts: densely connected layers, multi-scale extractor, multi-level feature fusion and residual learning. Densely connected layers result in a continuous memory (CM) mechanism. A contiguous memory mechanism is implemented by passing the state of the previous FSFblock to each layer of the current FSFblock. Different from previous experiments, we construct a three-bypass network and different bypasses use different convolutional kernels. In this way, the information flow among those bypasses can be shared with each other, which allow MLRN to detect the image features at different scales. The operation can be defined as:

$$F_{d,1} =[\, C_{3\times3}(F_{d-1})\, ,\, F_{d-1}] \tag{6}$$

$$F_{d,2} =[\, C_{3\times5}(F_{d-1})\, ,\, F_{d,1}] \tag{7}$$

$$F_{d,3} =[\, C_{5\times5}(F_{d-1})\, ,\, F_{d,2}] \tag{8}$$

$$F_d = F_{d,3} + F_{d-1} \tag{9}$$

where $C_{s\times s}$ means the S scale feature extractor. Our proposed the S scale feature extractor consist of two convolution layers of $s \times s$ kernel size and ReLU intermediate activation layer. The operation of $[\,\cdot\,]$ means the concatenation and $1 \times 1$ convolution, which is mainly designed for quickly fuse feature and reduce computational burden.

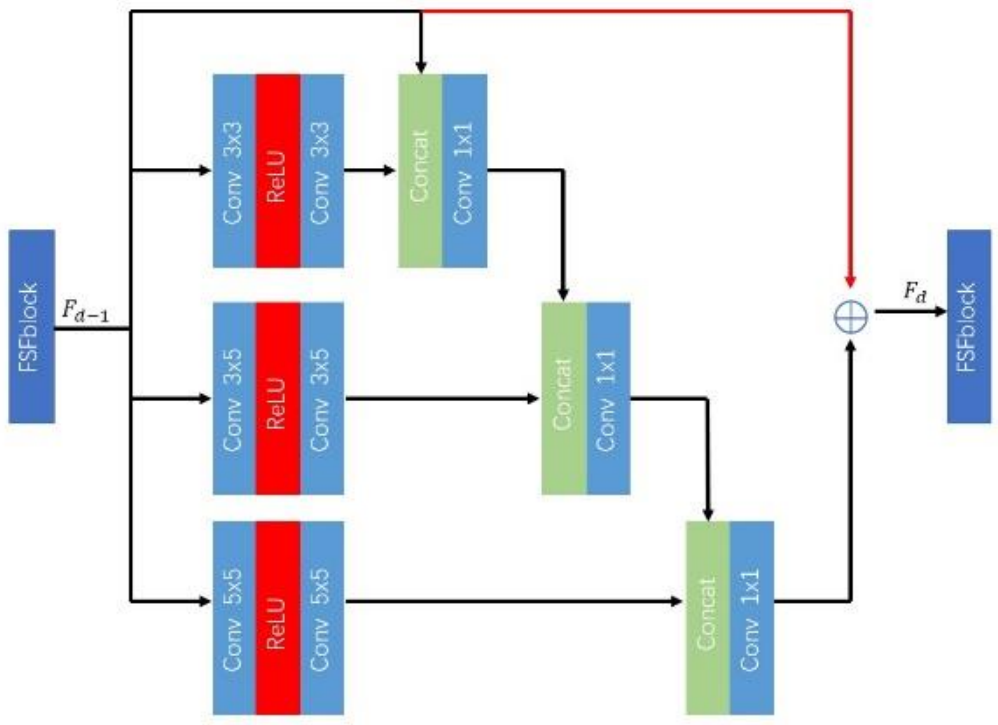

Fig. 3. Feature skip fusion block (FSFblock), which consists of multi-scale convolution layers and multi-level feature fusion for extracting multi-scale feature and fuse multi-level feature. Red arrows and black arrows mean residual information flow and global information flow respectively.

## D. *Implementation Details*

In our MLRN, we set the number of feedforward features G that remains at 32 and we use the zero padding to all convolution layers in order to remain the feature size unchanged. The number of FSFblock in MLRN is set to N = 8. Finally, MLRN outputs a RGB color three-channel image and can also process grayscale images. With regard to the loss function, we use L1 loss instead of L2 loss. Although L2 loss function is traditionally used, L1 loss have demonstrated faster convergence and improved performance.

## III. Discussions

### A. Difference to DenseNet

DenseNet [11] builds dense connections in any two layers of dense blocks [11]. However, such densely connected structures are only applied into a local manner because different dense blocks have different the size of the features so that dense blocks are unlikely to extract the raw features from subsequent features. Apart from that, the batch normalization (BN) layer was removed from our MLRN, which adds computational complexity and can't improve performance. In order to fix the feature size in the network unchanged, MLRN not use the pooling layer. What's more, feature skip fusion blocks are used to extract multi-scale features from all previous blocks and learn to fuse high-order features, leading to a contiguous memory mechanism that DenseNet [11] cannot implement.

### B. Difference to SRDenseNet

The SRDenseNet [12] have the same architecture as DenseNet [11]. SRDenseNet [12] introduced a dense block with a dense skip connection to solve the SISR. Although dense blocks in SRDenseNet can extract features in the block while constructing local skip connection for local residual learning. The block cannot directly extract the raw global feature, as our MLRN does. Global feature fusion and multi-level feature fusion is proposed in MLRN, and each FSFblock can extract from all global features of the previous block and local multi-scale feature. Taking full advantage of the global features and multi-scale feature, MLRN achieves better performance than SRDenseNet [12].

### C. Difference to MemNet

The difference between MemNet [8] and MLRN can be summarized into two points. First, the bottleneck layer fuse the global features with the 1×1 convolution layer. Multi-scale kernel size of convolutional layers are used in the feature skip fusion block (FSFblock). FSFblock can not only learn extracting multi-scale feature, as well as feature spatio-temporal correlation. Therefore, our FSFblock can further use a more global way than memory blocks to fuse intermediate features. Second, MemNet [8] does not extract features directly from LR images; it must adjust the size of the LR image by interpolation preprocessing to obtain the target size of the HR image, while our MLRN extracts original LR image features and directly utilizes sub-pixel convolution layer to reconstruct HR images.

## IV. Experiments

### A. Datasets and Metrics

Timofte et al. released a public open high-quality data set DVI2K [13] with a resolution of 2K. [13] Used for model training. DVI2K [13] can split 800 training images data, 100 validation images data and 100 test images data including different types of landscape images such as animals, humans, insects, buildings, plants and complex textures. The low-resolution images used for training were obtained by bicubic downsampling of ×2, ×3, ×4, are obtained using the MATLAB function with a bicubic function in 800 training images. We use benchmark data set, Set5 [14], Set14 [15], B100 [16], Urban100 [17] for test. For comparison between MLRN with other state of the art models, we use PSNR and SSIM [18] to evaluate as SISR results with different three scale factors.

### B. Training Details

As for training, we used 16 HR RGB image patches with a size of 192 × 192 randomly cropped from the training images in each training batch. The corresponding LR image for all models with different scale factors (×2, ×3 and ×4) becomes downsampling by adopting the MATLAB function with a bicubic function. During the training process, random vertical flip, random horizontal flip, and $90^{\circ}$ rotation augmented patches with a random probability of 0.5. Patch image pixel values are normalize and the average RGB values of the DIV2K [13] dataset which are subtracted from them as pre-processing. We use the Pytorch framework to implement our MLRN and train the model by setting up the ADAM optimizer [19] $\beta_1$= 0.9, $\beta_2$= 0.999, and $\varepsilon$=$10^{-8}$. The training loss function is the L1 loss. The learning rate for all layers is initialized to $1e^{-4}$, halving every 200 epochs, and 1,000 iterations of back-propagation constitute an epoch. Models with different scale factors will be trained from scratch. GPU GTX1080Ti takes about 5 day to train MLRN 1000 epochs.

### C. Ablation Study

Table 1 shows the ablation study that global feature fusion (GFF) and residual skip connection (RSC) how enhance performance for our model. We use four networks that set the same numbers of feed-forward features in FSFblock to build the basic model. The model N_BASE is obtained by removing GFF and RSC that is based on the basic MLRN, which has the standard framework. The performance (PSNR = 36.32 dB) of N_BASE is bad that is caused by the hard of training. It demonstrates that does not obtain better performance by stacking several basic convolution layers. Then, we add RSC and GFF to N_BASE to produce N_RSC and N_GFF. The results demonstrate each structure have the ability effectively improving the performance of N_BASE. This is largely due to the flow of gradients information enhanced by each structure. The combination of GFF and RSC will better than using it alone. When we use GFF and RSC at the same time (expressed as N_GFF_RSC), MLRN with RSC and GFF is clearly the best.

**Table 1.** Ablation study on effects of global feature fusion (GFF) and residual skip connection. We present the best performance (average PSNR) on Set5 with scale factor ×2 in 1000 epochs.

| | N_BASE | N_GFF | N_RSC | N_GFF_RSC |
|---|---|---|---|---|
| GFF | × | ✓ | × | ✓ |
| RSC | × | × | ✓ | ✓ |
| PSNR | 36.32 | 36.78 | 36.54 | 37.90 |

The visualization of the convergence process is shown in Fig 4. The curve verifies the analysis above, indicating that RSC and GFF further improve performance by stabilizing the training process convergence and accelerating training. As seen the red curve of N_GFF_RSC, we can see that RSC effectively improve performance while combined with GFF. Quantitative analysis shows that MLRN can benefit from RSC and GFF.

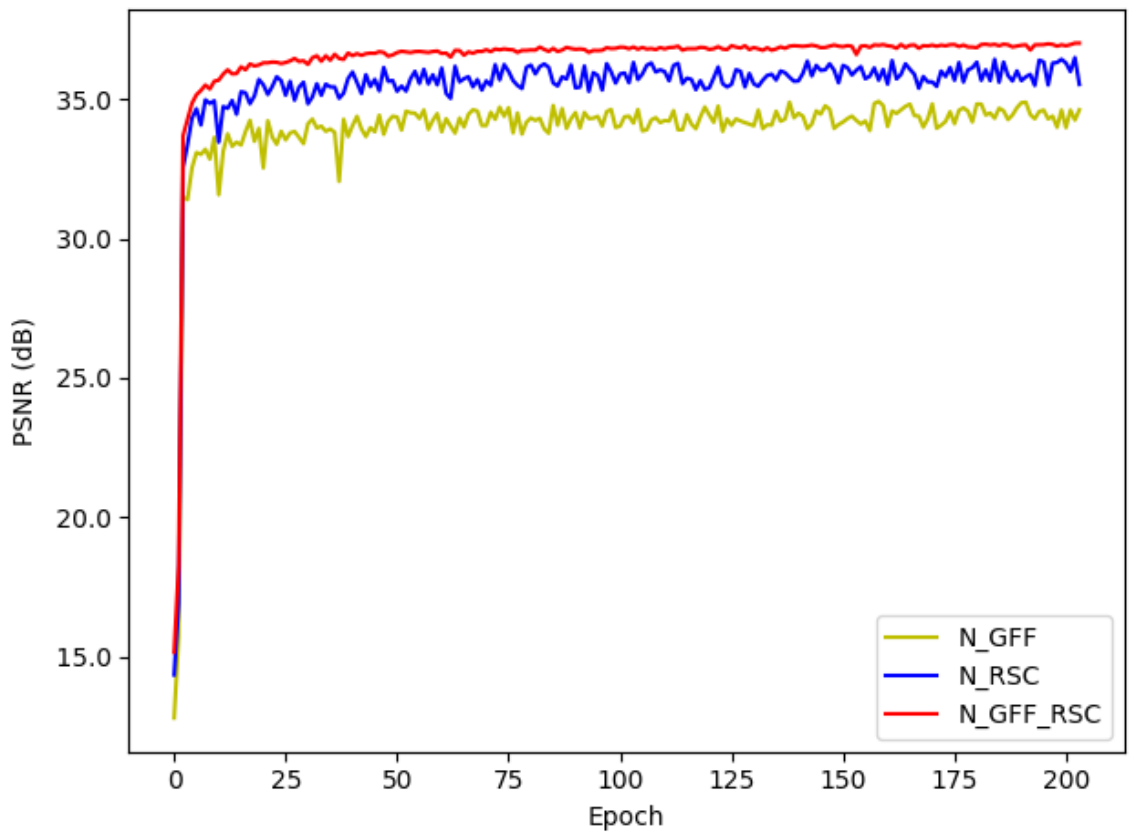

Fig. 4. Model convergence process model of different combinations of GFF and RSC. The curve for each structure is based on the average PSNR tested on DIV2K 890th-900th images at ×4 scale factor in 200 epochs. Other settings for the model are the same as described in this section.

### D. *Benchmark Results*

We also compare various SR methods between MLRN with other benchmarking methods, including Bicubic, SRCNN [4], VDSR [7], LapSRN [20], and MemNet [8]. We give the quantitative results of ×2, ×3, and ×4 in Table 2. Compared with persistent models such as MemNet [8], our MLRN performs best in all scaling factor benchmarks. When the scaling factor becomes larger (e.g. ×3, ×4), it becomes more hard for these models to recovery high resolution images from low resolution images with much lower clue since need more texture details to reconstruct. Our MLRN is still superior to others. Although MemNet [8] can fuse global information to recovery similar structures by gate units, MLRN also better than it. This indicates feature skip fusion block (FSFblock) is more effective than the memory block of MemNet [8] and further indicates that the integration of global intermediate features through global feature fusion (GFF) providing more clues to recovery high quality image from low quality image data. Compared to other state of the art methods, our MLRN can still get the best average results across all scale and all datasets.

**Table 2.** Public benchmark test results. Average PSNR/SSIMs for scale factor ×2, ×3, and ×4 on datasets Set5, Set14, BSD100, and Urban100.

| Dataset | Scale | Set5 | Set14 | BSD100 | Urban100 |
|---|---|---|---|---|---|
| Bicubic | × 2 | 33.66/0.9299 | 30.24/0.8688 | 29.56/0.8431 | 26.88/0.8403 |
| | × 3 | 30.39/0.8682 | 27.55/0.7742[1] | 27.21/0.7382 | 24.46/0.7349 |
| | × 4 | 28.42/0.8104 | 26.00/0.7027 | 25.96/0.6675 | 23.14/0.6577 |
| SRCNN | × 2 | 36.66/0.9542 | 32.42/0.9063 | 31.36/0.8879 | 29.50/0.8946 |
| | × 3 | 32.75/0.9090 | 29.28/0.8208 | 28.41/0.7863 | 26.24/0.7989 |
| | × 4 | 30.48/0.8628 | 27.49/0.7503 | 26.90/0.7101 | 24.52/0.7221 |
| VDSR | × 2 | 37.53/0.9587 | 33.03/0.9124 | 31.90/0.8960 | 30.76/0.9140 |
| | × 3 | 33.66/0.9213 | 29.77/0.8314 | 28.82/0.7976 | 27.14/0.8279 |
| | × 4 | 31.35/0.8838 | 28.01/0.7674 | 27.29/7251 | 25.18/0.7524 |
| LapSRN | × 2 | 37.52/0.9591 | 33.08/0.9130 | 30.41/0.9101 | 37.27/0.9740 |
| | × 3 | 33.82/0.9227 | 29.79/0.8320 | 27.07/0.8272 | 32.19/0.9334 |
| | × 4 | 31.51/0.8855 | 28.19/0.7720 | 25.21/0.7553 | 29.09/0.8893 |
| MemNet | × 2 | 37.78/0.9597 | 33.28/0.9142 | 32.08/0.8978 | 31.31/0.9195 |
| | × 3 | 34.09/0.9248 | 30.00/0.8350 | 28.96/0.8001 | 27.56/0.8376 |
| | × 4 | 31.74/0.8893 | 28.26/0.7723 | 27.40/0.7281 | 25.50/0.7630 |
| MLRN | × 2 | 37.90/0.9601 | 33.49/0.9140 | 32.11/0.8989 | 31.87/0.9260 |
| | × 3 | 34.18/0.9254 | 30.22/0.8369 | 29.01/0.8033 | 27.88/0.8469 |
| | × 4 | 31.92/0.8911 | 28.43/0.7748 | 27.49/0.7334 | 25.78/0.7763 |

Visual comparison of scale factor ×4 are shown in Fig.5. Most of the comparison methods, like SRCNN and VDSR, reconstruct visible and blurred textures, and some textures and edges cannot be recovered. In contrast, our MLRN reconstruct sharper edges and textures with fewer artifacts than other models, closer to the original image (HR). Other methods cannot successfully reconstruct it, and our MLRN can use it to restore its edges sharply sharper. This is mainly because our MLRN make the best of the global feature with global feature fusion and multi-scale feature with multi-level feature fusion.

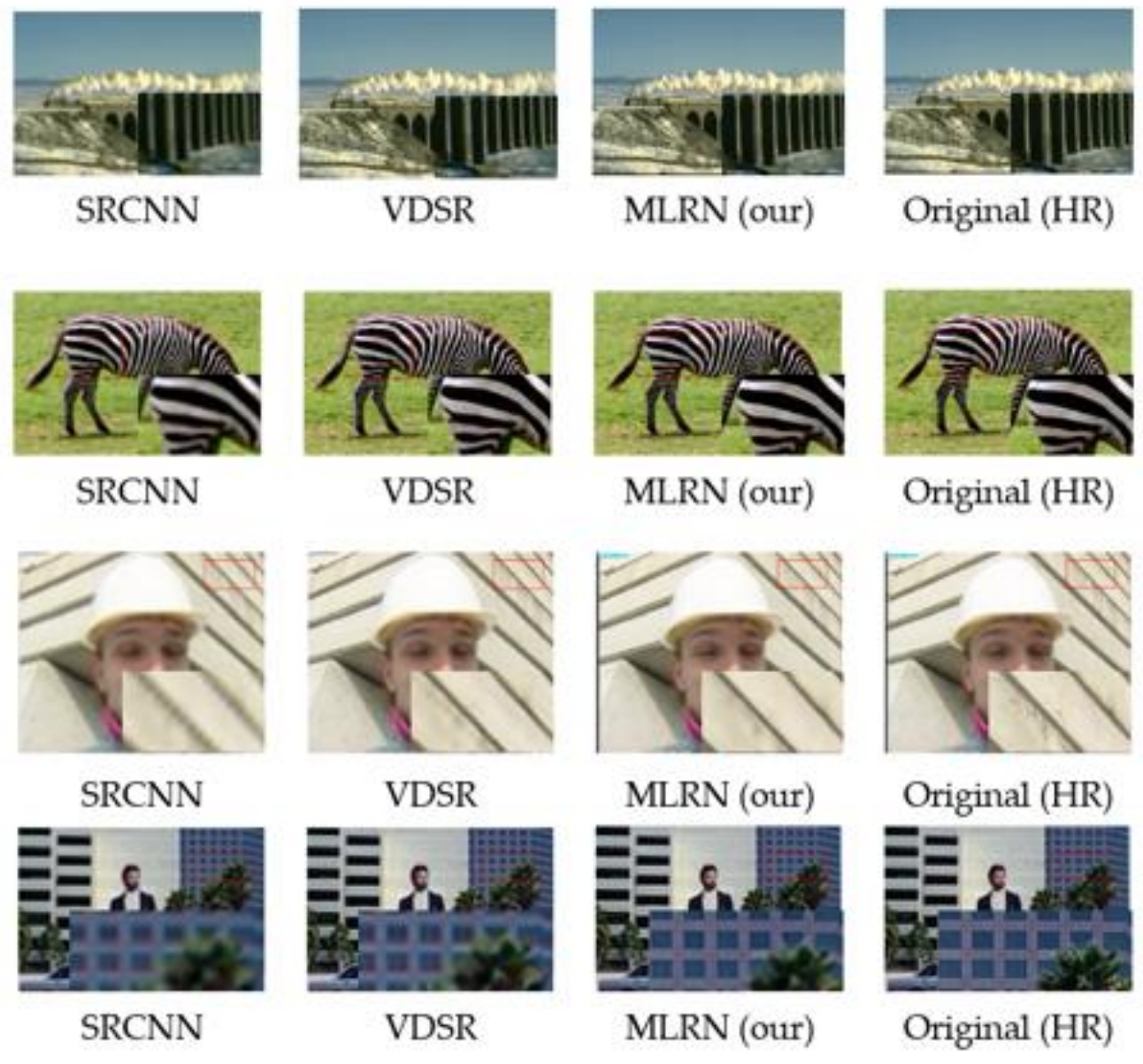

Fig. 5. Visual comparison of MLRN with other SR methods on ×4 scale SR task. Obviously, MLRN can reconstruct realistic images with sharp edges.

## V. CONCLUSIONS

We propose a super-resolution model named Multi-Level Feature Fusion network (MLRN) for SISR in which a feature skip fusion block (FSFblock) is proposed to build a standard module. Each FSFblock can extract the multi-scale feature and fuse multi-level feature directly from all previous blocks in MLRN. The global feature fusion (GFF) in the bottleneck block further constructs a dense feature fusion for the FSFblock resulting in improving the global information flow and stabilizing the training process. Gradient leads to a global continuous information memory mechanism. In addition, our MLRN can directly extracts raw features from the original low-quality image with dense features and directly reconstructs high quality images. There is no image scaling preprocessing. By making the best use of global feature and local multi-scale feature, our MLRN can perform a wide range of networks. Quantitative and visual evaluation have proved our MLRN has better performance over the most advanced models.

## REFERENCES

[1] X. Zhang and X. Wu, "Image interpolation by 2-D autoregressive modeling and soft-decision estimation," IEEE Trans. Image Process., vol. 17, no. 6, pp. 887-896.

[2] G. Freedman and R. Fattal, "Image and video upscaling from local self-examples," ACM Trans. Graph., vol. 30, no. 11, pp. 12, 2011.

[3] J.Yang,J. Wright, T. S. Huang, andY. Ma,"Imagesuper-resolution via sparse representation," IEEE Trans. Image Process., vol. 19, no. 11, pp. 2861–2873, Nov. 2010.

[4] C. Dong, C. C. Loy, K. He, and X. Tang, "Learning a deep convolutional network for image super-resolution," European Conference on Computer Vision., pp. 184–199, 2014.

[5] C. Dong, C. C. Loy, K. He, and X. Tang, "Accelerating the superresolution convolutional neural network," European Conference on Computer Vision., pp. 391–407, 2016.

[6] W. Shi, J. Caballero, F. Huszar, J. Totz, A. Aitken, R. Bishop, D. Rueckert, and Z. Wang, "Real-time single image and video super-resolution using an efficient sub-pixel convolutional neural network," Proceedings of the CVPR 2016.,pp.27-30, June 2016.

[7] J. Kim, J. K. Lee, K. M. Lee, "Accurate image super-resolution using very deep convolutional networks," Proc. IEEE Conf. Comput. Vis. Pattern Recognit., pp. 1646-1654, Jun. 2016.

[8] Y. Tai, J. Yang, X. Liu, C. Xu, "MemNet: A persistent memory network for image restoration, " Proc. IEEE Int. Conf. Comput. Vis., pp. 4539-4547, Aug. 2017.

[9] M. Haris, G. Shakhnarovich, N. Ukita, "Deep backprojection networks for super-resolution", Proc. Conf. Comput. Vis. Pattern Recognit., pp. 1664-1673, 2018.

[10] Y. Zhang, Y. Tian, Y. Kong, B. Zhong, Y. Fu, "Residual dense network for image super-resolution," Proc. IEEE Conf. Comput. Vis. Pattern Recognit., pp. 2472-2481, Jun. 2018.

[11] G. Huang, Z. Liu, "Densely connected convolutional networks," Proc. Conf. Comput. Vis. Pattern Recognit., pp. 1-4, 2017.

[12] R. Timofte, E. Agustsson, L. Van Gool, M.-H. Yang, L. Zhang, B. Lim, S. Son, H. Kim, S. Nah, K. M. Lee, "NTIRE 2017 challenge on single image super-resolution: Methods and results," Proc. IEEE Conf. Comput. Vis. Pattern Recognit. Workshops, pp. 1110-1121, 2017.

[13] R. Zeyde, M. Elad, M. Protter, "On single image scale-up using sparse-representations", Proc. 7th Int. Conf. Curves Surfaces, pp. 711-730, 2012.

[14] D. Martin, C. Fowlkes, D. Tal, J. Malik, "A Database of Human Segmented Natural Images and Its Application to Evaluating Segmentation Algorithms and Measuring Ecological Statistics," Proc. IEEE Int'l Conf. Computer Vision, 2001-July.

[15] J. B, A. Singh, N. Ahuja, "Single image super-resolution from transformed self-exemplars," Proc. IEEE Conf. Comput. Vis. Pattern Recog., pp. 5197-5206, 2015.

[16] Zhou Wang, A. C. Bovik, H. R. Sheikh and E. P. Simoncelli, "Image quality assessment: from error visibility to structural similarity," in IEEE Transactions on Image Processing, vol. 13, no. 4, pp. 600-612, April 2004.

[17] D. Kingma, J. Ba, Adam: A method for stochastic optimization, 2014.

[18] W.-S. Lai, J.-B. Huang, N. Ahuja, M.-H. Yang, "Deep Laplacian pyramid networks for fast and accurate super-resolution," Proc. IEEE Conf. Comput. Vis. Pattern Recognit., pp. 5835-5843, 2017.